\documentclass[12pt]{article}
\usepackage[top=1in, bottom=1in, left=1in, right=1in]{geometry}
\usepackage{geometry}
\usepackage{amsthm, soul, amsfonts}
\usepackage{amsmath, amssymb}
\usepackage{authblk, array, setspace} 
\usepackage{natbib}
\usepackage[pdftex,colorlinks=true,linkcolor=blue,citecolor=blue,urlcolor=blue,bookmarks=false,pdfpagemode=None]{hyperref}
\usepackage{enumerate}
\usepackage{graphicx}
\usepackage{float}
\usepackage{pgffor}
\newtheorem{theorem}{Theorem}[section]
\newtheorem{lemma}[theorem]{Lemma}

\newtheorem{assumption}{Assumption}
\usepackage{hyperref}
\usepackage{algorithm}
\usepackage{algorithmic}

\newcommand{\imagefolder}{pics/}

\author{Nicolo Colombo\footnote{\tt nicolo.colombo@rhul.ac.uk} and 
Yang Gao\footnote{\tt yang.gao@rhul.ac.uk} \\
  Department of Computer Science\\
  Royal Holloway University of London,
  Egham Hill, Egham TW20 0EX, UK 
 }

\title{Disentangling Neural Architectures and Weights: \\ 
A Case Study in Supervised Classification}
\usepackage{xcolor}

\begin{document}

\maketitle

\begin{abstract}
The history of deep learning has shown that 
human-designed problem-specific networks can greatly 
improve the classification performance of general neural models.
In most practical cases, however, choosing the optimal 
architecture for a given task remains a challenging problem. 
Recent architecture-search methods 
are able to automatically build neural models with 
strong performance but fail 
to fully appreciate 
the interaction between neural architecture and weights.

This work investigates  
the problem of disentangling 
the role of the neural structure and its edge weights,
by showing that 
well-trained architectures may not 
need any link-specific fine-tuning of the weights. 
We compare the performance of such weight-free 
networks (in our case these are binary networks with 
\{0, 1\}-valued weights) with random, 
weight-agnostic, pruned and   
standard fully connected networks.
To find the optimal weight-agnostic network, we use  
a novel and computationally efficient method that translates 
the hard architecture-search problem into a feasible 
optimization problem.
More specifically, we look at the optimal task-specific architectures 
as the optimal configuration of binary 
networks with \{0, 1\}-valued 
weights, which can be found through an approximate gradient 
descent strategy. 
Theoretical convergence guarantees of the proposed algorithm are 
obtained by bounding the error in the gradient approximation and 
its practical performance is evaluated 
on two real-world data sets.
For measuring the structural similarities between different 
architectures, we use a novel spectral 
approach that allows us to underline the intrinsic differences between real-valued networks and weight-free architectures.

\end{abstract}

\section{Introduction}
The exceptionally good performance of 
neural-based models can be considered the main responsible 
for the recent huge success of AI.
To obtain the impressive learning and predictive power
of nowadays models, 
computer scientists have spent years designing, 
fine-tuning and trying 
different and increasingly sophisticated 
network architectures.
In particular, it has become clear that the 
optimal edge structure greatly 
depends on the specific task to be solved, 
e.g.
CNNs for image classification
\cite{szegedy2017inception,wei2015hcp},
RNNs  for sequence labeling \cite{ma2016end,graves2012supervised} and 
Transformers for language modeling
\cite{vaswani2017attention,devlin2019bert}.
But this has been a pretty painful path as 
designing new neural architectures 
often requires considerable human effort and 
testing them is a challenging computational task. 

Existing methods for a fully-automated architecture search 
are mostly based on two strategies:
(i) \emph{architecture search} schemes   \cite{zoph2016neural,liu2018darts,gaier2019weight,you2020greedynas}, 
which
look for optimal neural structures in a given search space 
\footnote{Usually, the boundary of the search spaces are set 
by limiting the number of allowed neural operations, 
e.g. node or edge addition or removal.}, and  
(ii) \emph{weight pruning} procedures
\cite{Han2015LearningBW,Frankle2019TheLT,Zhou2019DeconstructingLT}, 
which attempt to improve the performance of large 
(over-parameterized) networks by  removing 
the `less important' connections.
Neural models obtained through these methods 
have obtained good results 
on benchmark data sets for 
image classification (e.g. CIFAR-10) or 
language modeling (e.g. Penn Treebank) \cite{yu2019playing}. 
What remains unclear is whether 
such networks perform well because of 
their edge structures, 
their optimized weight values or a 
combination of the two.
A puzzling example can be found in the framework of 
\emph{weight-agnostic} neural networks \cite{gaier2019weight},
where models are trained and tested by 
enforcing all connections to share the same weight value.
Their decent classification power 
suggests that the contribution of fine-tuned weights 
is limited \cite{gaier2019weight} 
but the conclusion should probably be revised 
as recent studies \cite{zhang2020deeper} show that 
the output can actually be very sensitive 
to the specific value of the shared weight. 

In this work, we try to disentangle the role
of neural architecture and weight values by considering
a class of \emph{weight-free} neural models 
which do not 
need any tuning or random-sampling of the 
connections weight(s).
To train such models in a weight-free fashion, 
we define a new algorithm that optimizes 
their edge structure directly, 
i.e.\ without 
averaging over randomly-sampled weights, 
as for weight-agnostic networks \cite{gaier2019weight},  
or training edge-specific real-value weights, 
as in other architecture
search methods \cite{liu2018darts,Frankle2019TheLT}.
The proposed scheme is obtained by formulating 
the architecture search problem 
as an optimization task over the space of 
all possible binary networks with \{0, 1\}-valued weights.
Given a fully connected neural model with connections 
${\cal C} = \{ i\}_{i=1}^d$, for example,
the search space is the power set ${\mathcal P}({\cal C})$ 
and the goal is 
to find a subset of 
nonzero connections ${\cal C}_* \subset {\cal C}$ 
that minimizes a given 
objective function.\footnote{Not surprisingly, the size of the search space is 
exponentially large, as
$|{\mathcal P}({\cal C})| = 2^d$, 
which reflects the 
hardness of the original discrete optimization problem.} 
%
In practice, formulating architecture search 
as a discrete optimization problem is 
advantageous because
i) the neural architecture associated with 
the optimal binary-weight optimal assignment, 
${\cal C}^* \subset {\cal C}$, does not depend on 
specific weight values and
ii) the discrete optimization problems can be approximated 
arbitrarily well by continuous optimization problems.
\begin{figure}[t]
    \centering
    \includegraphics[width=0.9\textwidth]{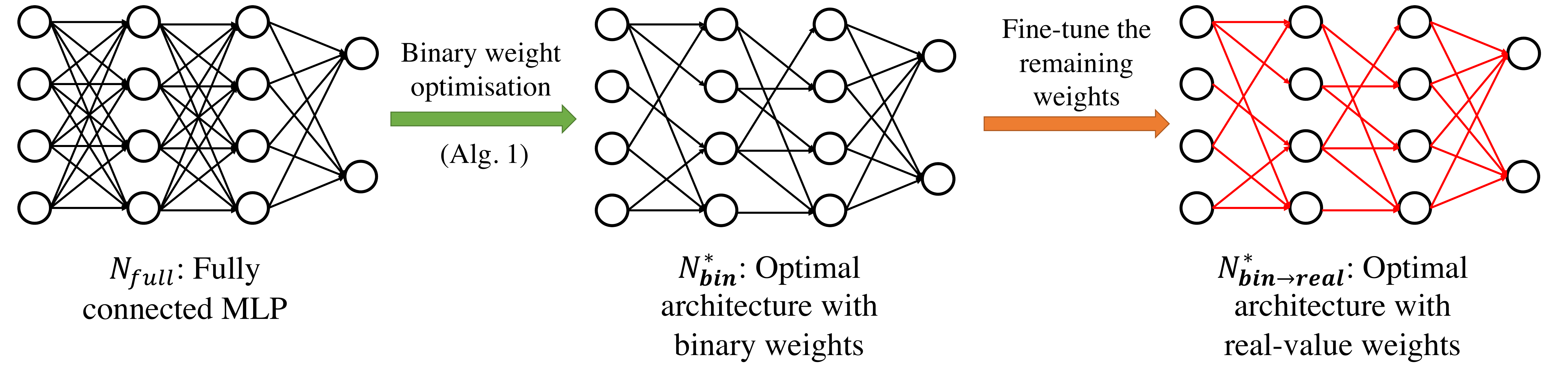}
    \caption{We first obtain the optimal neural 
    architecture $N^*_{\textbf{bin}}$ by performing 
    binary weight optimization on 
    the base network $N_{\mbox{full}}$, 
    and then fine tune real-value weights on 
    top of $N^*_{\textbf{bin}}$ to obtain $N^*_{\textbf{bin} \to \textbf{real}}$.
    The performance gap between $N^*_{\textbf{bin}}$
    and $N^*_{\textbf{bin} \to \textbf{real}}$ 
    is the contribution of the weight values.}
    \label{fig:overview}
\end{figure}

To analyze the intrinsic predictive power 
of weight-free networks, 
we compare their performance with a real-valued version 
of them, where fine-tuned weights are attached to all 
non-vanishing connections, i.e. a real-valued network 
with weights $\{w_i > 0\}_{i \in {\cal C_*}}$.
Figure~\ref{fig:overview} illustrates the major steps of our 
strategy.
Finally, we summarize the structural differences of 
the obtained networks, by looking at the spectrum 
of their \emph{Laplacian matrix}.
Graph spectral analysis has been used in computational 
biology for comparing the structure of \emph{physical} neural networks, 
e.g. the brain \emph{connectome}
\cite{de2014laplacian,villegas2014frustrated,4390947},
but, to the best of our knowledge, rarely exploited to evaluate or classify 
\emph{artificial} neural models in an 
AI or machine learning framework.

The contributions of this work are the following:
\begin{itemize}
\item 
We formulate the architecture search task 
as a binary optimization problem and propose
a computationally efficient algorithm.
\item
We provide theoretical convergence guarantees
for the proposed method.
To the best of our knowledge, this is the first time 
convergence guarantees have been provided for an 
architecture-search method.
\item 
We quantitatively measure the contribution 
of neural architectures and weight by comparing the 
performance of weight-free and real-valued 
networks on image and text classification tasks. 
Results suggest that binary networks trained through 
the proposed algorithm often 
outperform both binary networks obtained through other 
techniques and real-valued ones (probably because 
they are less prone to training data over-fitting).
\item 
We propose a novel and bilogically-inspired method 
to measure the similarity
between neural architectures based on the eigen-spectrum of  
their Laplacian matrix \cite{de2014laplacian}. 
\end{itemize}

\section{Related Work}
{\bf Neural architecture-search} methods
construct a neural architecture incrementally,
by augmenting a connection or an operation at each step.
Early methods 
\cite{stanley2002evolving,zoph2016neural,real2019regularized} 
employ expensive \emph{genetic algorithms} 
or \emph{reinforcement learning}
(RL). 
%
More recent schemes
either design differentiable losses
\cite{liu2018darts,xie2019snas},
or use random weights to 
evaluate the performance of the architecture 
on a validation set 
\cite{gaier2019weight,pham2018efficient}. 
%
Unlike these methods, that search architectures by
adding new components, our method removes redundant 
connections from  
an over-parameterized `starting' network,
e.g.\ a fully-connected multi-layer
perceptron (MLP). 
%
%
{\bf Network pruning} approaches start from pre-trained 
neural models and prune the unimportant connections 
to reduce the model size and achieve better performance 
\cite{han2015deep,collins2014memory,Han2015LearningBW,Frankle2019TheLT,yu2019playing}.
These methods prune 
existing task-specific network architectures 
e.g.\ 
CNNs for image classification, 
whereas our method starts from 
generic architectures (that would 
perform poorly without pruning),
e.g.\ fully-connected MLPs.
In addition, our method does not require any 
pre-training.
%
%
{\bf Network quantization and binarization}
reduces the computational
cost of neural models by using lower-precision 
weights \cite{jacob2018quantization,zhou2016dorefa}, 
or mapping and hashing similar
weights to the same value 
\cite{chen2015compressing,hu2018hashing}.
In an extreme case, the weights, 
and sometimes even the inputs, are binarized,
with
positive/negative weights 
mapped to $\pm1$
\cite{soudry2014expectation,courbariaux2016binarized,hubara2017quantized,shen2019searching,courbariaux2015binaryconnect}.
%
%
As a result, these 
methods keep all original connections, 
i.e. do not perform any architecture search.
%



\section{Background}
\paragraph{Supervised learning of real-valued models}
Let  $\mathcal{D} = 
\{z_i = (x_i, y_i) \in {\cal X} \times {\cal Y}\}_{i = 1}^n$
be a training set 
of vectorized objects, $x_i$, and labels, $y_i$.
Let 
$f: {\cal W}\times {\cal X} \times {\cal Y} \to \mathbf{R}$,
where ${\cal W}$ is a given \emph{parameter space},
be such that $f(w, z)$ yields the \emph{miss-classification 
value} associated with object-label pair $z \in {\cal D}$.
Furthermore, let $f$ be of the form 
$f(w, z) = \ell(\psi(w, z), y)$, where 
$\ell:{\cal Y} \times {\cal Y} \to {\mathbf R}$ is 
a fixed \emph{cost function} and
$\psi:{\cal W} \times {\cal X} \to {\cal Y}$
the \emph{classification model}.
For any fixed choice of parameter $w \in {\cal W}$, $\psi(w, z)$ maps vectorized objects $x \in {\cal X}$ 
into labels $y \in {\cal Y}$.
Classifiers such as $\psi$ can be trained 
by minimizing the average loss
\begin{align}
\label{eq:std_loss}
F(w) = |\mathcal{D}|^{-1}
\sum_{z \in {\cal D}} f(w, z).
\end{align}
In standard neural training 
${\cal W} = {\mathbf R}^d$
and 
$w_* = \arg \min_{w \in {\cal W}}F(w)$ 
is usually found 
through Stochastic Gradient Descent (SGD) updates 
\begin{align}
\label{eq:std_sgd}
w^t \leftarrow w^{t-1} - \frac{\eta^t}{|\mathcal{Z}^t|}  \sum_{z \in \mathcal{Z}^t}  
\frac{\partial f(z; w^{t-1})}{\partial w^{t-1}}, 
\end{align}
where $\eta^t > 0$ is the learning rate
and $\mathcal{Z}^t \subseteq \mathcal{D}$ a randomly selected 
mini-batch at step $t$.

\paragraph{Training binary networks}
Unlike conventional neural models, where 
${\cal W} = {\mathbf R}^d$,
a binary-weight neural model is a function 
$\psi: {\cal W} \times {\cal X} \to {\cal Y}$,
where ${\cal W} = \{a, b\}^d$, 
with $a, b\in {\mathbf R}$ being fixed values.
Here we focus on a special class of 
binary models where $a = 0$ and $b = 1$ but other choices 
are possible, e.g. a popular setting is $a = -1$ and $b = 1$.

Similarly to the case ${\cal W} = {\mathbf R}^d$, 
the single-input 
miss-classification value is 
$f:\{0, 1\}^d \times {\cal X} \times {\cal Y} \to {\mathbf R}$
and $\psi$ can be trained by setting 
$W  =  \{0, 1\}^d$ \eqref{eq:std_loss}.
Minimizing \eqref{eq:std_loss}
over $\{0, 1 \}^d$, however, is 
a challenging discrete optimization problem, with the 
size of the search space, ${\mathbf P}(\{ i\}_{i=1}^d)$, 
increasing exponentially with the growth of $d$.

\paragraph{SGD for binary optimization}
One possible way to minimize \eqref{eq:std_loss} 
over ${\cal W} = \{0, 1\}^d$
is to 
i) define a 
suitable \emph{continuous approximation} of 
the minimization problem 
$\min_{w \in \{ 0, 1\}^d} F(w)$, 
ii) solve such a continuous approximation 
through standard SGD updates and 
iii) binarize the obtained solution at the end.
The strategy proposed here is to consider an approximate 
binary-weight constraints, $\{0, 1 \} \to [0, 1]$
and minimize \eqref{eq:std_loss} over ${\cal V} = [0, 1]^d$
under the further (but more handable) constraint 
$v  = B_{M}(w)$, where   
$B_{M}: {\mathbf R}^d \to {\cal V}$, $w \in {\mathbf R}^d$ 
is an \emph{approximate binarization function} 
and  $M > 0 $ a parameter controlling the 
`sharpness' of the approximation. 
More explicitly, we propose to train $\psi$
by solving 
\begin{align}
\label{eq:bin_loss real}
\begin{array}{ll} 
    \text{minimize} & F(B_M(w)) \\
    \text{s.t.} &  w \in {\mathbf R}^d 
\end{array}
\quad  \Leftrightarrow \quad 
\begin{array}{ll} 
    \text{minimize} & F(v) \\
    \text{s.t.} & v = B_{M}(w) \text{ and  } 
    w \in {\mathbf R}^d 
\end{array}
\end{align}
where $F$ is given in \eqref{eq:std_loss}
and $B_M:{\mathbf R}^d \to [0, 1]^d$ is the approximate 
step function\footnote{
The original problem is recovered for 
$M = \infty$, as $B_{\infty}(w)$ is 1  
if $w\geq 0$ and 0 otherwise.
}
\begin{align*}
B_M(w)= \sigma(M w) = (1+ \exp(-M\cdot w))^{-1}  
\end{align*}
As $B_M$ is differentiable for all $M < \infty$,  
we can solve (\ref{eq:bin_loss real})
through SGD updates 
\begin{align}
\label{eq:bin_sgd}
w^t & \leftarrow  
w^{t-1} - 
\frac{\eta^t}{|\mathcal{Z}^t|}  \sum_{z \in \mathcal{Z}^t}  
\nabla f(v^{t-1}, z) \circ \sigma_M'(w),
\end{align}
where $[\nabla f(v, z)]_i = \frac{\partial f(v, z)}{\partial v_i}$,  
`$\circ$' denotes element-wise vector multiplication and 
$\sigma_M'(w) = \sigma_M(w) \circ (1 - \sigma_M(w)) = B_M(w)\circ (1 - B_{M}(w))$ is the 
first derivative of $B_M(w)$\footnote{
The extra factor in \ref{eq:bin_sgd} (compared to \eqref{eq:std_sgd}) 
takes into account the composition $F(v) = F(B_M(u))$.
}.

\section{Method}
\label{sec:method}

\begin{figure}[H]
    \begin{center}
    \includegraphics[scale=0.9]{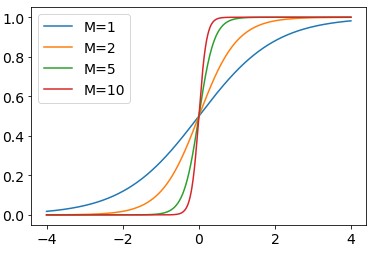} 
    \includegraphics[scale=0.9]{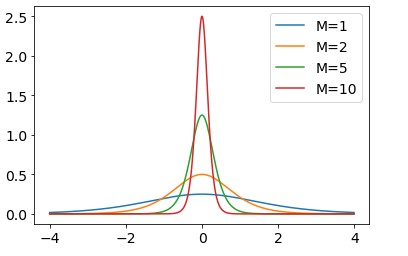} 
    \caption{Shape of the parameterized binarization function $B_M$ (upper) and its derivative (bottom) with different values of $M$ . \label{fig:sigma}}
    \end{center}
\end{figure}

A major difficulty of solving \eqref{eq:bin_loss real}
is that, as $M$ in $B_M$ increases, the approximation
quality improves but gradient-based
updates may become exponentially small.
Figure \ref{fig:sigma} shows the shape of $B_{M}(w)$ and its first derivative,  
$B_M'(w) = \nabla B_M(w) = B_M(w)\circ (1 - B_M(w))$, 
for different values of $M$.
In particular, note that 
$B_M'(w) \approx 0$ in large parts of its domain, even 
for reasonably small values of $M$.
Inspired by the BinaryConnect approach \cite{courbariaux2015binaryconnect},
we define two binarization functions (by setting $M$ to two 
different values):
i) $B_{M_{hard}}$ with $M_{hard} >> 1$ , to be used 
at forward-propagation time, and 
ii) $B_{M_{soft}}$, with $M_{soft} < M_{hard}$, to compute the gradient updates.
The idea is to solve  
\eqref{eq:bin_loss real} with $M = M_{hard}$, 
which is a good approximation of the original
discrete optimization problem, 
through the relaxed SGD weight updates:\footnote{
Note that $w \in {\mathbf R}^d$ is now a completely 
\emph{unconstrained} optimization variable.} 
\begin{align}
\label{eq:soft_bin_sgd}
w^t \leftarrow w^{t-1} - 
\frac{\eta^t}{|\mathcal{Z}^t|} 
\sum_{z \in \mathcal{Z}^t} 
\nabla f(B_{M_{hard}}(w^{t - 1}), z) \circ B_{M_{soft}}'(w^{t-1})
\end{align}
Algorithm \ref{alg:bin} is a pseudo-code
implementation of this idea.


\begin{algorithm}[H]
\caption{SGD-based architecture search (one epoch)
\label{alg:bin}
}
\textbf{Inputs:} \\
\hspace*{5pt} Randomly initialized 
neural model with real-valued weights, 
$w^0 \in \mathbf{R}^d$,
training data set,
$\{z_i, \cdots, z_n\}$, 
learning rates, $\{\eta^1, \cdots, \eta^{T}\}$,
binarization function, $B_M$, 
soft- and hard-binarization constants, $M_{hard} >> 1$ and 
$M_{soft} < M_{hard}$ \\
\\
\textbf{Binary optimization by SGD:}
\begin{algorithmic}
\FOR{$t = 1, \cdots, T$}
    \STATE Obtain mini-batch: $\mathcal{Z} = \{z_i, \cdots, z_{i+k-1}\}$
    \STATE Compute hard-binarized weights: 
    $v = B_{M_{hard}}(w^{t-1})$
    \STATE Compute loss: 
    $F(v) = k^{-1} \sum_{z \in  \mathcal{Z}} f(v, z)$
    \STATE Update unconstrained weights: 
    $w^t \leftarrow w^{t-1} - \eta^{t}  k^{-1} 
    \sum_{z \in \mathcal{Z}} 
    \nabla f(v, z) \circ B_{M_{soft}}'(w^{t-1})$
\ENDFOR
\end{algorithmic}
\textbf{\\Extract architecture:}
\begin{algorithmic}
\STATE Obtain binary network 
$\psi_{bin} 
= \lim_{M \to \infty} \psi(B_M(w^T), z)$ 
\end{algorithmic}
\end{algorithm}

\begin{figure}[t]
    \begin{center}
	\includegraphics[width=.99\textwidth]{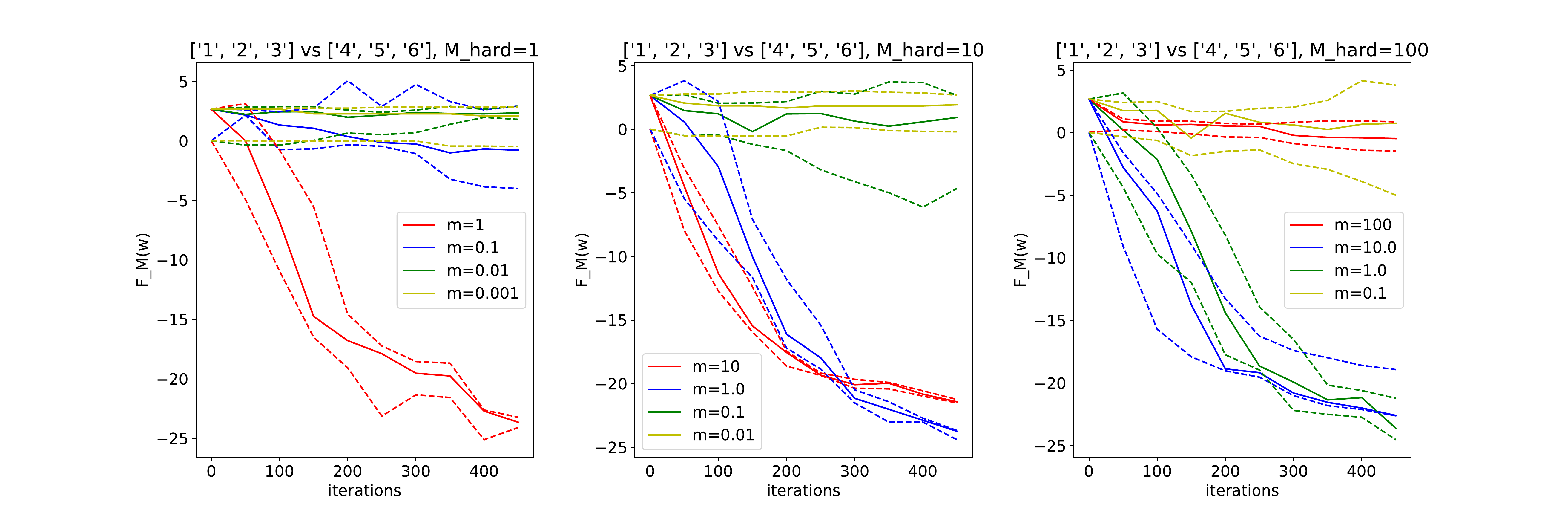}
    \caption{
    Convergence of Algorithm \ref{alg:bin} for varying values of 
    $M_{hard}$ and $M_{soft}$. 
    The learning task is to 
	    distinguish between two groups of hand-written digits, 
	    $\{1, 2, 3\}$ and $\{ 4, 5,6 \}$. 
	    We use 
    a constant learning parameter and 
    100 images of each class as training set 
    (see Section \ref{section experiments} for
    detailed experimental setup).
	Solid and dashed lines correspond to the median and the 25th/75th percentiles over 10 independent runs.
	}
    \label{fig:convergence}
\end{center}
\end{figure}

\paragraph{Remarks}
The hard-binarization parameter, 
$M_{hard}$, which defines the optimization problem 
to be solved, should be fixed a priori (we 
set $M_{hard} = 50$ in the experiments of 
Section \ref{section experiments}).
The soft-binarization parameter, 
$M_{soft}$, which defined the slackness of 
the gradient approximations, could 
in principle be tuned through standard cross-validation procedures,
but we keep it fixed in this work ($M_{soft} = M_{hard} / 10$ 
in Section \ref{section experiments}).
Figure \ref{fig:sigma} shows 
the quality of the continuous approximation for different values of 
$M_{hard}$. 
Large values of $M_{hard}$ make the 
optimization task harder but reduce the error made in
converting the approximately-binary trained network, 
$\psi(B_{M_{hard}}(w_*), z)$ with $B_{M_{hard}}(w_*) \in [0, 1]$ 
and $w_* = w^T$ in Algorithm \ref{alg:bin}, 
to the `true' binary network  
$\psi_{bin}(w, z) = \lim_{M \to \infty} \psi(B_M(w_*), z)$ 
(see last line of Algorithm \ref{alg:bin}). 
Figure \ref{fig:convergence}
shows the convergence of 
Algorithm \ref{alg:bin} for different choices of $M_{hard}$ and 
$M_{soft}$.
Note that choosing $M_{soft} << M_{hard}$ 
may be counterproductive when $M_{hard}$ is small
but consistently helps the SGD updates converge 
when $M_{hard} >> 1$.\footnote{Intuitively, this 
is because too-small values of $M_{soft}$ make 
\eqref{eq:soft_bin_sgd} a very bad approximation 
of the true SGD updates, i.e. \eqref{eq:soft_bin_sgd} 
with ($M_{soft} = M_{hard}$).}

\paragraph{Algorithm convergence}
In the remaining of this section, we consider 
single-input mini-batch, i.e. we set ${\cal Z}^t  = \{ \tilde z\}$ 
for all $t =1, \dots, T$ and 
prove that the  Algorithm \ref{alg:bin} 
converges to a local optimum of 
\eqref{eq:bin_loss real}.
As a consequence, we can state that Algorithm \ref{alg:bin} 
can be consistently used to solve arbitrarily good approximations
of the unfeasible discrete optimization problem 
``$\min_{w \in \{0,1\}^d} F(w)$'' with $F$ given in 
\eqref{eq:std_loss}.
We show that the approximate gradient updates 
\eqref{eq:soft_bin_sgd} are 
close-enough to the gradient updates needed for 
solving \eqref{eq:bin_loss real}, i.e. the 
sharp ($M = M_{hard}$) approximation of
the original discrete problem.

We first assume that the loss
function $f$ is differentiable over 
${\cal V} = [0, 1]^d$ and  
\begin{align}
	\label{assumptions}
	\max_{v \in {\cal V}, z \in {\cal X} \times {\cal Y}} 
	\| \nabla f(v, z) \|^2 \leq G^2, \quad \mbox{and}
	\quad
	f(v, z) - f(v', z) \geq \nabla f(v', z)^T (v - v')
\end{align}
and prove the following lemma. 
\footnote{Due to the space limit, proofs for all
lemmas and theorems are put 
to the Supplementary Materials.}
\begin{lemma}
	\label{lemma big F}
	Assume that $f$ 
	meets the requirements
	in \eqref{assumptions}, then $F(v): [0, 1]^d \to {\mathbf R}$  
	defined as 
	$F(v) = |{\cal D}|^{-1} \sum_{z \in {\cal D}} f(v, z)$
	is also a convex function, i.e. 
	$ F(v) - F(v') \geq \nabla F(v')^T (v - v')$
\end{lemma}
\paragraph{Remarks}
The convergence of the stochastic gradient descent does not 
follow automatically from  
the assumption on $f$  and Lemma \ref{lemma big F} for two reasons:
i) $F(v) = |{\cal D}|^{-1} \sum_{z \in {\cal D}} f(v, z)$
is component-wise composed with the non-convex function 
$B_{M_{hard}}(w)$  and
ii) we use an approximation of the `true'  
gradient of $F(B_{M_{hard}}(w))$.

To show that the approximated updates \eqref{eq:soft_bin_sgd}
can be used to solve \eqref{eq:bin_loss real} we need  
the following lemma. 
\begin{lemma}
	\label{lemma error}
	    For any $w^t$, $t = 0, \dots, T$, and $\tilde z \in {\cal D}$,
	    the gradient updates \eqref{eq:soft_bin_sgd} are 
	    equivalent to 
	   \begin{align}
	\label{updates v}
	v^{t+1} &\leftarrow v^t - \eta_t (\nabla f(v^t, \tilde z) + r^t), 
    \end{align}
    where $v^t = B_{M_{hard}}(w^t)$, 
    \begin{align}
    \label{eq:rewrite}
    r^{t}& = \nabla f(v^t, \tilde z)  
- B_{M_{hard}}'(\xi^t) \circ \nabla f(v^t,\tilde z) 
\circ B_{M_{soft}}'(w^t), 
\end{align}
with $\xi^t \in [w^t, w^t - \eta_t \nabla f(v^t, \tilde z) 
\circ B_{M_{soft}}(w^t)]$.  
	Furthermore, the error terms, $r^t$, $t = 1, \dots, T$, obey 
	\begin{align}
		\| r^t \|^2 \leq G^2 C^2 
		\quad C = 1 + \frac{M_{hard} M_{soft}}{16} 
	\end{align}
with $G$ defined in (\ref{assumptions}).
\end{lemma}
\paragraph{Remark} Lemma \ref{lemma error} is 
essentially a consequence of the Mean Value Theorem. 
The proof starts by applying $B_{M_{hard}}$ to both sides of 
\eqref{eq:soft_bin_sgd} and then uses 
$f(a + b) = f(a) + \nabla f(c)^T b$, $c \in [a, a + b]$ 
to isolate the first term in the 
right-hand side of \eqref{eq:rewrite}.

Finally, 
based on the lemmas above and an additional
assumption that $\eta_t = \frac{c}{t}$ for all $t=1, \dots, T$, $c \in \mathbf{R}_+$, we obtain the following theorem, which
proves that the approximate updates \eqref{updates v}
converge to $w_* = \text{arg } \min_{w \in {\mathbf R}} F(B_{M_{hard}}(w))$.

\begin{theorem}
	\label{theorem convergence}
	Assume that the cost function $f: [0, 1]^d \to {\mathbf R}$ 
	meet the requirements 
	in \eqref{assumptions}, and let 
	$F(w) = |{\cal D}|^{-1} 
	\sum_{z \in{\cal D}} f(B_{M_{hard}}(w), z)$,
	$w^T$ be 
	the output of Algorithm \ref{alg:bin}, 
	$w^* = \text{arg }\min_{w \in {\mathbf R}^d} 
	F(B_{M_{hard}}(w))$,  
	then 
	\begin{align}
		\mathbf{E}\left(F(B_{M_{hard}}(w^T)) - F(B_{M_{hard}}(w^*))\right) 
		\leq 
		G^2 \frac{c \left(1 + C^2\right)}{2} \frac{1 + \log T}{T},
	\end{align}
	where $\mathbf{E}(\cdot)$ is the expectation over the 
	random sampling of $\tilde z \in {\cal D}$ and $C$ is the constant defined 
	in Lemma \ref{lemma error}.
\end{theorem}

\section{Experiments}
\label{section experiments}
\paragraph{Data sets}
We run a series of similar training-testing   
experiments on two real-world 
data sets, ${\cal D} \in \{{\cal D}_{mnist}, 
{\cal D}_{citeseer} \}$.
The first is
a collection of \emph{vectorized images} of hand-written 
digits from the MNIST database\footnote{We use the 
csv files from \url{https://www.kaggle.com/oddrationale/mnist-in-csv}.} 
and the second 
a collection of \emph{vectorized abstracts} of 
scientific papers from the \emph{Citeseer} database.\footnote{Data available at \url{http://networkrepository.com/citeseer.php}. 
 }
%
Both data sets consist of labeled objects that we  
represent through $(I=64)$-dimensional,
real-valued, unit norm 
vectors, i.e. $|x| = 64$ and $x^Tx  = 1$ for all 
$x \in {\cal D}_{u}$, $u \in \{ mnist, citeseer \}$. 
The original MNIST images are resized (by cropping 2 pixels 
per side and pooling over 64 non-overlapping windows), 
vectorized and normalized. 
The original \emph{bag-of-words}, 
3703-dimensional, integer-valued  embedding vectors\footnote{
See `readme.txt' in the 
Citeseer dataset for more 
details of the pre-processing steps performed.
}
 of 
the Citeseer abstracts are reduced to $I$ 
dimensions by projecting 
into the $d$-dimensional space spanned by the first $I$ 
Principal Components of the entire corpus.

\paragraph{Models}
For all experiments, the architecture 
search-space boundaries are defined by  
\begin{align}
	\label{psi}
	\psi(w, x) = \text{ReLU}\left( 
	\sum_{i=1}^I \tanh\left( \sum_{j=1}^I 
	[w_2]_{ij} \tanh\left(\sum_{k=1}^I[w_1]_{jk} x_k
	\right)
	\right)
	\right)
\end{align}
where $x \in {\cal X} = {\mathbf R}^{I}$, 
$[w_1, w_2] \in {\cal W}$, 
${\cal W} = \{0, 1\}^{d}$ for the binary networks 
and ${\cal W} = {\mathbf R}_+^{d}$ for the 
real networks\footnote{For all real-valued networks, 
we enforce nonnegativity by letting 
$w = u^2$, $u \in {\mathbf R}^d$. 
Preliminary simulations (data not shown) suggest that, in 
most cases, 
the constraint does not compromise significantly the model performance.
}, $d = 2 I^2$, 
$\tanh(w)$ is the hyperbolic tangent activation function 
and $\text{ReLU}(w) = \log(1 + e^{w})$.
In all cases, the classifier $\psi$ in (\ref{psi}) is trained by 
minimizing 
\begin{align}	
	\label{exp objective}
	\quad F(w)= |{\cal D}|^{-1}\sum_{(x, y) \in {\cal D}} 
	\psi(w, x)  (1 - 2 y) 
\end{align}
over ${\cal W} \in \{ \{0, 1 \}^d, {\mathbf R}^d_+ \}$. \footnote{
Note that \eqref{exp objective} is 
equivalent to \eqref{eq:std_loss} where 
$f(y, y') = y' (1 - 2 y)$
and $y' = \psi(w, x)$.} 
The nonnegativity constraint imposed on 
real-networks is not essential but 
we introduce it here to make the  
comparison between real- and $\{0, 1\}$-valued networks 
fairer and conceptually easier.  
We compare five different training strategies: 

\begin{itemize}
\item 
{\bf real} and {\bf real$\to$bin}: 
the real network is $\psi(w, x)$ with
$w = \min_{w \in {\mathbf R}^d_+} F(w)$
and 
the binary network $\psi(w',x)$ with 
\begin{align}
	\label{choose threshold}
	w' = B_{\infty}(w - \xi_*), \quad
	\xi_* = \text{arg }\min_{\xi \in T_w} 
	F(B_{\infty}(w - \xi))
\end{align}
where $T_w$ is the set of the 
$\{10, \cdots, 100\}$th
percentiles of all entries of $w$.
\item 
{\bf lottery} and {\bf lottery$\to$bin}:
the real network is $\psi(w, z)$ 
with 
$$w = \arg 
\min_{1\leq t\leq T} F(w^t)$$ 
where the elements 
of the sequence 
$w^1, \dots, w^T $,  
$w^t \in {\mathbf R}_+^d$ are defined recursively 
by   $w^{1} = \min_{w \in {\mathbf R}^d_+} F(w)$
and $w^t = w^{t-1} \circ w'$, with $w'$ obtained as 
in (\ref{choose threshold}) with $w = w^{t - 1}$,
and 
the binary network is $\psi(w', x)$ with 
$w' = w \circ \tilde w$, with $\tilde w$ 
obtained as in (\ref{choose threshold}) 
(see \cite{Frankle2019TheLT} for 
more details about this procedure). 
\item 
{\bf bin} and {\bf bin$\to$real}: 
the binary network is $\psi(w', z)$ with 
$$w' = \min_{w \in \{0, 1\}^d}F(w)$$
obtained from Algorithm \ref{alg:bin}
and the real network is $\psi(w, z)$ where 
$$w = \text{arg } 
\min_{w \in {\mathbf R}_+^d} F(w \circ w')$$
\item 
{\bf random} and {\bf random$\to$bin}:
the real network is $\psi(w, x)$ with 
$w  = u^2$, $u\sim {\cal N}(0, 1)^{d}$
and the binary network is $\psi(w', x)$ with 
$w'$ obtained as in 
(\ref{choose threshold}).
\footnote{This optimization step may partially explain
why {\bf random$\to$bin} 
performs consistently better than  {\bf random}, which 
is purely untrained.}
\item 
{\bf agnostic} and {\bf agnostic$\to$real}: 
the binary network is $\psi(w', x)$ with 
$$w' = 
\arg\min_{\bar w \in S} 
\left(\min_{\tilde w \in R_{\bar w} } 
F(\tilde w)\right)$$ 
where $\bar w \in S$ are random architectures 
$\bar w = B_{\infty}(w^t - \xi)$,
$\xi \in T_w$ 
defined as in (\ref{choose threshold}),    
$w^t = u^2$, 
$u \sim {\cal N}(0, 1)^d$, 
$t = 1, \dots, T$, 
and $\tilde w \in R_{\bar w}$ are 
shared-weight 
networks defined by 
$\tilde w = u_0 \bar w$, 
$u_0 \sim {\cal N}(0, 1)$, and 
the real network is $\psi(w, x)$ 
with $w  = u_0 \bar w'$, $u_0 \sim {\cal N}(0, 1)$.
\end{itemize}

\paragraph{Results}

\begin{figure}
\begin{center}
\begin{tabular}{ccc}
	\includegraphics[trim=60 80 80 80, clip=1, width=.43\textwidth]{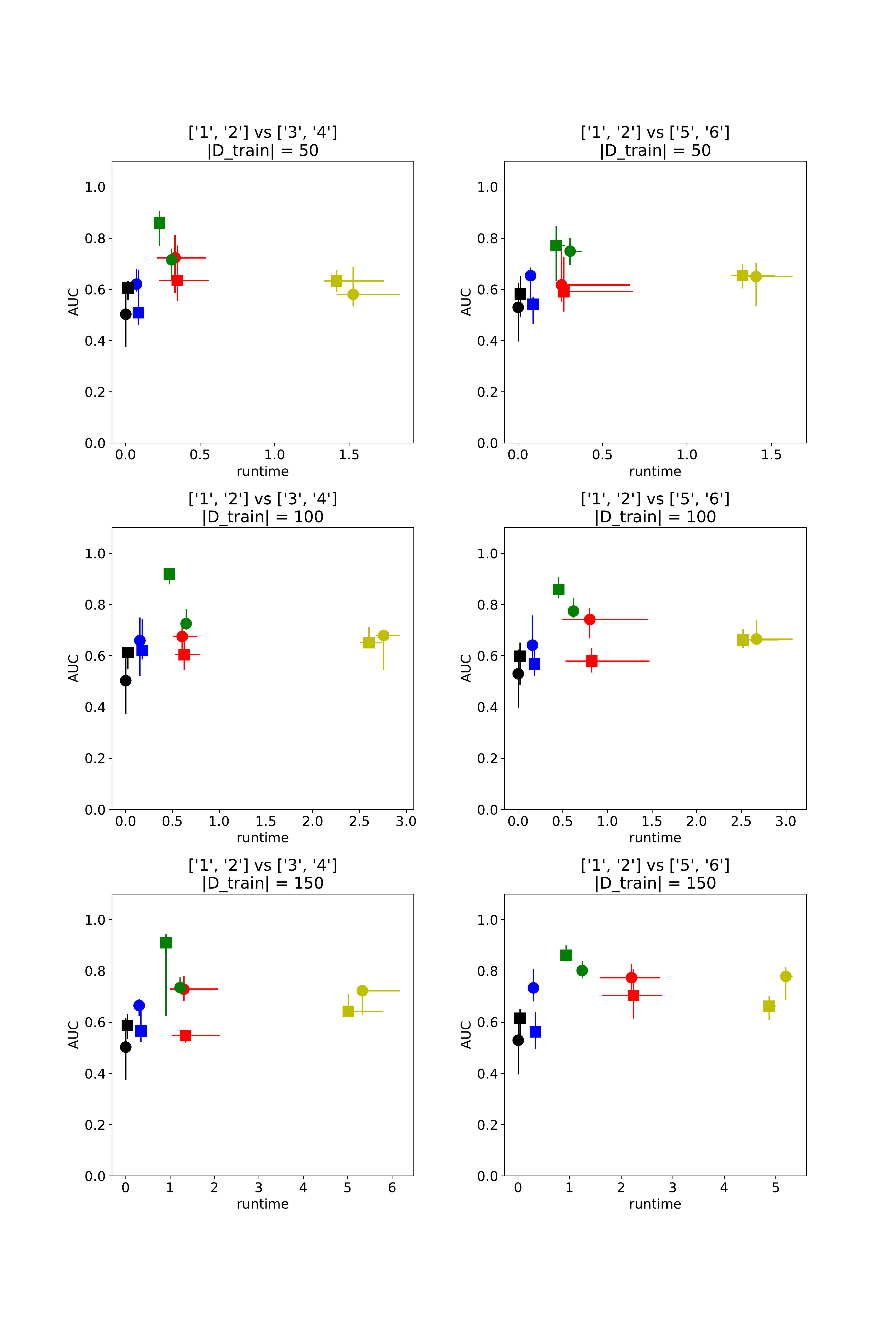} 
	&
	\includegraphics[trim=80 80 60 80, clip=1, width=.43\textwidth]{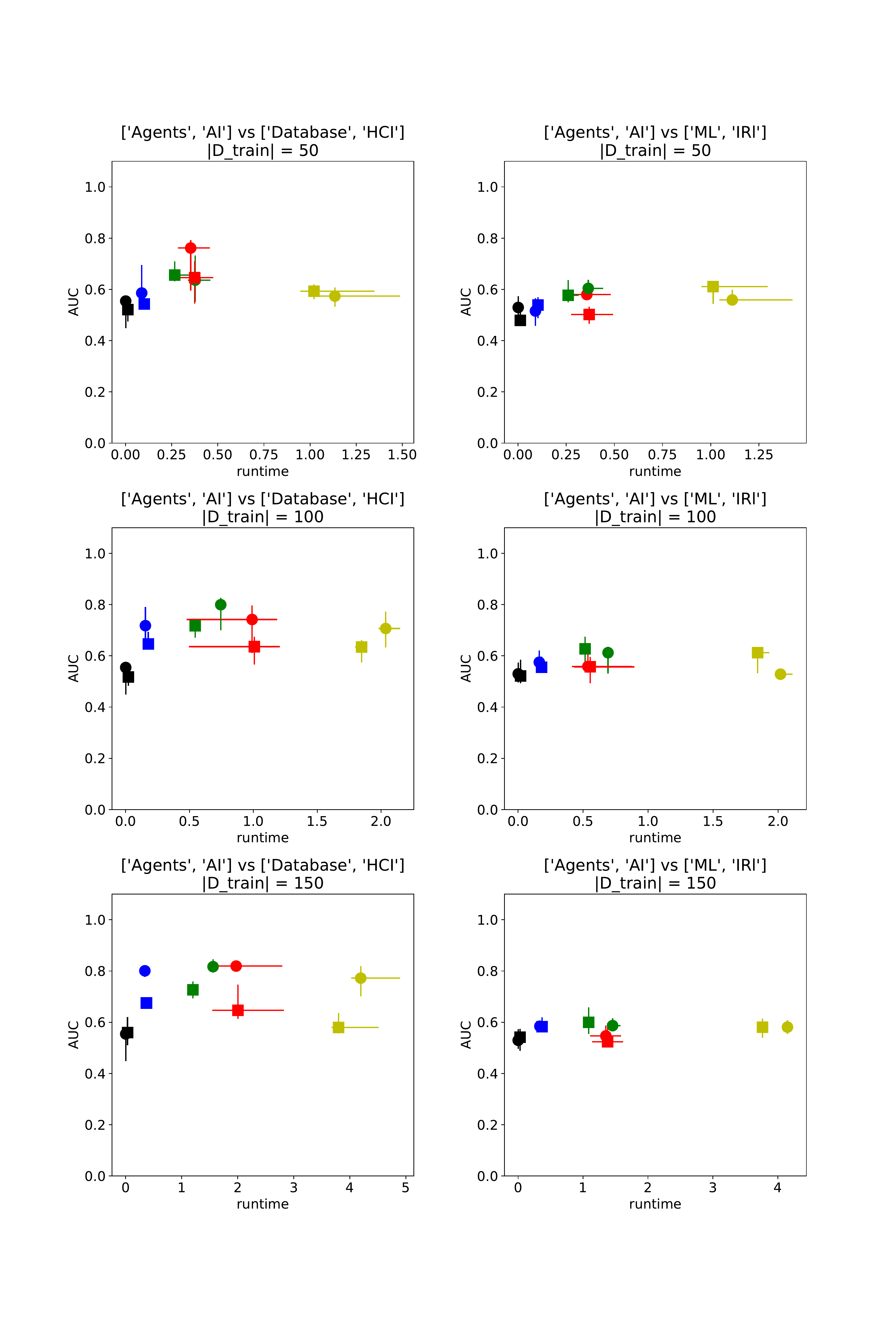}
	&
	\includegraphics[trim=30 0 0 0, clip=1,width=.07\textwidth]{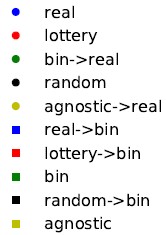}
\end{tabular}
	\caption{Runtime vs. AUC  (over 50 test images) of the models 
	trained to solve different tasks with training 
	data sets of different sizes from the MNIST data set (columns 1 and 2) 
	and the Citeseer data set (columns 3 and 4).
	}
\label{fig:results}
\end{center}
\end{figure}
Figure \ref{fig:results}
shows the performance of all methods listed above. 
For each data set, we train the models to distinguish between the two (non-overlapping) 
\emph{groups of classes} reported by the plot titles.
The scatter plots show the median and 
quantiles (error bars) 
of the run time over 10 independent experiments 
(x-axis)
and AUC scores of the models (y-axis).
For each task, we use a test set of 50 objects 
per class and three training sets of different sizes.
The proposed model, {\bf bin} or {\bf bin$\to$real} (in green), achieves the
best performance with 
comparable run time in all but one cases.
In particular, it is striking to see that 
weight-free networks may so often 
outperform fine-tuned more flexible models.
From an architecture-optimization perspective, 
the proposed method seems to produce 
better networks than both  
weight-agnostic search 
(agnostic, in yellow)
and pruning (lottery ticket, 
in red) methods.

\paragraph{Laplacian spectra}
\begin{figure}
\begin{center}	
	\includegraphics[trim=20 0 0 20, clip=0,width=.87\textwidth]{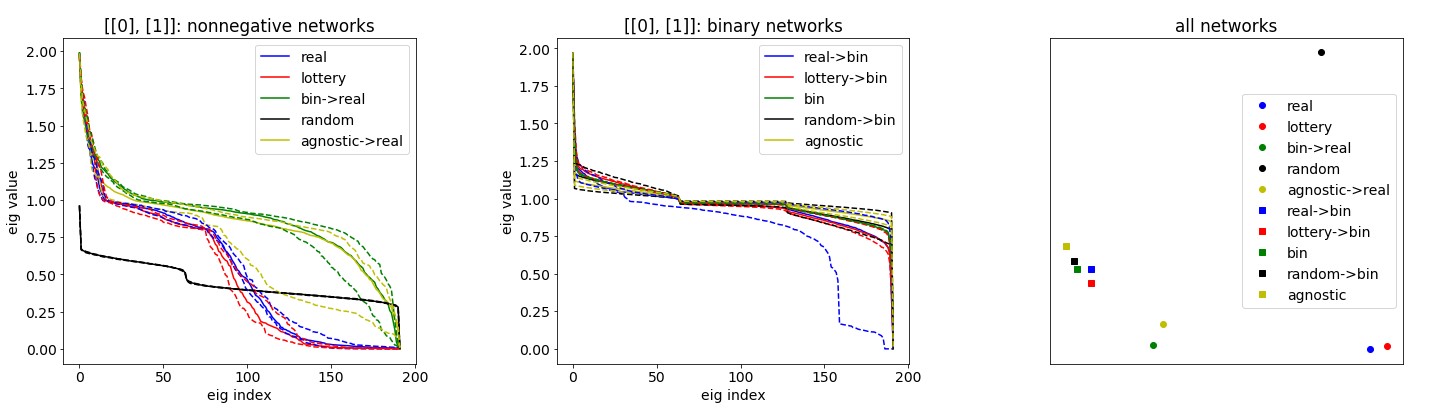}
	\label{figure spectra}
	\caption{Citeseer data: spectrum of the network normalized 
	Laplacian matrices (plots 1 and 2) and a 2-dimensional 
	(PCA) representation of their distances. 
	}
\end{center}
\end{figure}
Figure \ref{figure spectra} shows the 
spectrum of the normalized Laplacian matrix  
of real and binary networks trained for discriminating 
images of digits 1 and 2 from the
MNIST data set.\footnote{Given $w = W_1,W_2$, with 
$W_i = \text{mat}(w_i) \in {\mathbf R}_+^{\tilde d \times \tilde d}$, $\tilde d = \sqrt{d / 2}$, the Laplacian matrix is computed from the block adjacency matrix $A = [A_1, A_2, A_3]^T$ with $A_1= [1, W_2, 0]$, $A_2 = [W_2^T, 1, W_1]$ and $A_3 = [W_1^T 0, 1]$.} 
Solid and dashed lines in the first two plots 
correspond to the median and quartiles of the
eigenvalues (ordered by magnitude) obtained in 10 
independent runs. 
The last plot is a 2-dimensional 
reduction (PCA) of the 
$3I$-dimensional vector space associated 
with the Laplacian spectra.
Distances between different markers 
can be seen as a representation of the 
structural differences between models.\footnote{
Note that the spectral representation
solves automatically any (unavoidable) 
hidden nodes relabeling ambiguity.}

\section{Discussions}

The experiments show that weight-free networks 
found by our method can
perform surprisingly well on different classification tasks
and even outperform more flexible
models. 
Real-valued fine-tuning of the edge structures 
may help when the classification task is hard but may also
cause performance drop 
due to data overfitting. 
An analysis of the obtained networks based on their 
Laplacian spectra shows that different training strategies lead to highly different real-valued models but `spectrally similar'
architectures. 
An interesting question is whether such spectral similarities 
may be exploited in a transfer learning framework, and we 
leave this as future work.

\bibliography{refs}

\begin{thebibliography}{10}

\bibitem{chen2015compressing}
Wenlin Chen, James Wilson, Stephen Tyree, Kilian Weinberger, and Yixin Chen.
\newblock Compressing neural networks with the hashing trick.
\newblock In {\em International conference on machine learning}, pages
  2285--2294, 2015.

\bibitem{collins2014memory}
Maxwell~D Collins and Pushmeet Kohli.
\newblock Memory bounded deep convolutional networks.
\newblock {\em arXiv preprint arXiv:1412.1442}, 2014.

\bibitem{courbariaux2015binaryconnect}
Matthieu Courbariaux, Yoshua Bengio, and Jean-Pierre David.
\newblock {BinaryConnect}: Training deep neural networks with binary weights
  during propagations.
\newblock In {\em Advances in neural information processing systems}, pages
  3123--3131, 2015.

\bibitem{courbariaux2016binarized}
Matthieu Courbariaux, Itay Hubara, Daniel Soudry, Ran El-Yaniv, and Yoshua
  Bengio.
\newblock Binarized neural networks: Training deep neural networks with weights
  and activations constrained to+ 1 or-1.
\newblock {\em arXiv preprint arXiv:1602.02830}, 2016.

\bibitem{de2014laplacian}
Siemon de~Lange, Marcel de~Reus, and Martijn Van Den~Heuvel.
\newblock The laplacian spectrum of neural networks.
\newblock {\em Frontiers in computational neuroscience}, 7:189, 2014.

\bibitem{devlin2019bert}
Jacob Devlin, Ming-Wei Chang, Kenton Lee, and Kristina Toutanova.
\newblock Bert: Pre-training of deep bidirectional transformers for language
  understanding.
\newblock In {\em Proceedings of the 2019 Conference of the North American
  Chapter of the Association for Computational Linguistics: Human Language
  Technologies, Volume 1 (Long and Short Papers)}, pages 4171--4186, 2019.

\bibitem{Frankle2019TheLT}
Jonathan Frankle and Michael Carbin.
\newblock The lottery ticket hypothesis: Finding sparse, trainable neural
  networks.
\newblock {\em arXiv: Learning}, 2019.

\bibitem{gaier2019weight}
Adam Gaier and David Ha.
\newblock Weight agnostic neural networks.
\newblock In {\em Advances in Neural Information Processing Systems}, pages
  5365--5379, 2019.

\bibitem{graves2012supervised}
Alex Graves.
\newblock Supervised sequence labelling.
\newblock In {\em Supervised sequence labelling with recurrent neural
  networks}, pages 5--13. Springer, 2012.

\bibitem{han2015deep}
Song Han, Huizi Mao, and William~J Dally.
\newblock Deep compression: Compressing deep neural networks with pruning,
  trained quantization and huffman coding.
\newblock {\em arXiv preprint arXiv:1510.00149}, 2015.

\bibitem{Han2015LearningBW}
Song Han, Jeff Pool, John Tran, and William~J. Dally.
\newblock Learning both weights and connections for efficient neural network.
\newblock {\em ArXiv}, abs/1506.02626, 2015.

\bibitem{hu2018hashing}
Qinghao Hu, Peisong Wang, and Jian Cheng.
\newblock From hashing to cnns: Training binary weight networks via hashing.
\newblock In {\em Thirty-Second AAAI Conference on Artificial Intelligence},
  2018.

\bibitem{hubara2017quantized}
Itay Hubara, Matthieu Courbariaux, Daniel Soudry, Ran El-Yaniv, and Yoshua
  Bengio.
\newblock Quantized neural networks: Training neural networks with low
  precision weights and activations.
\newblock {\em The Journal of Machine Learning Research}, 18(1):6869--6898,
  2017.

\bibitem{jacob2018quantization}
Benoit Jacob, Skirmantas Kligys, Bo~Chen, Menglong Zhu, Matthew Tang, Andrew
  Howard, Hartwig Adam, and Dmitry Kalenichenko.
\newblock Quantization and training of neural networks for efficient
  integer-arithmetic-only inference.
\newblock In {\em Proceedings of the IEEE Conference on Computer Vision and
  Pattern Recognition}, pages 2704--2713, 2018.

\bibitem{liu2018darts}
Hanxiao Liu, Karen Simonyan, and Yiming Yang.
\newblock {DARTS}: Differentiable architecture search.
\newblock {\em arXiv preprint arXiv:1806.09055}, 2018.

\bibitem{ma2016end}
Xuezhe Ma and Eduard Hovy.
\newblock End-to-end sequence labeling via bi-directional lstm-cnns-crf.
\newblock In {\em Proceedings of the 54th Annual Meeting of the Association for
  Computational Linguistics (Volume 1: Long Papers)}, pages 1064--1074, 2016.

\bibitem{pham2018efficient}
Hieu Pham, Melody~Y Guan, Barret Zoph, Quoc~V Le, and Jeff Dean.
\newblock Efficient neural architecture search via parameter sharing.
\newblock {\em arXiv preprint arXiv:1802.03268}, 2018.

\bibitem{real2019regularized}
Esteban Real, Alok Aggarwal, Yanping Huang, and Quoc~V Le.
\newblock Regularized evolution for image classifier architecture search.
\newblock In {\em Proceedings of the AAAI conference on artificial
  intelligence}, volume~33, pages 4780--4789, 2019.

\bibitem{4390947}
M.~{Reuter}, M.~{Niethammer}, F.~{Wolter}, S.~{Bouix}, and M.~{Shenton}.
\newblock Global medical shape analysis using the volumetric laplace spectrum.
\newblock In {\em 2007 International Conference on Cyberworlds (CW'07)}, pages
  417--426, 2007.

\bibitem{shen2019searching}
Mingzhu Shen, Kai Han, Chunjing Xu, and Yunhe Wang.
\newblock Searching for accurate binary neural architectures.
\newblock In {\em Proceedings of the IEEE International Conference on Computer
  Vision Workshops}, pages 0--0, 2019.

\bibitem{soudry2014expectation}
Daniel Soudry, Itay Hubara, and Ron Meir.
\newblock Expectation backpropagation: Parameter-free training of multilayer
  neural networks with continuous or discrete weights.
\newblock In {\em Advances in Neural Information Processing Systems}, pages
  963--971, 2014.

\bibitem{stanley2002evolving}
Kenneth~O Stanley and Risto Miikkulainen.
\newblock Evolving neural networks through augmenting topologies.
\newblock {\em Evolutionary computation}, 10(2):99--127, 2002.

\bibitem{szegedy2017inception}
Christian Szegedy, Sergey Ioffe, Vincent Vanhoucke, and Alexander~A Alemi.
\newblock Inception-v4, inception-resnet and the impact of residual connections
  on learning.
\newblock In {\em Thirty-first AAAI conference on artificial intelligence},
  2017.

\bibitem{vaswani2017attention}
Ashish Vaswani, Noam Shazeer, Niki Parmar, Jakob Uszkoreit, Llion Jones,
  Aidan~N Gomez, {\L}ukasz Kaiser, and Illia Polosukhin.
\newblock Attention is all you need.
\newblock In {\em Advances in neural information processing systems}, pages
  5998--6008, 2017.

\bibitem{villegas2014frustrated}
Pablo Villegas, Paolo Moretti, and Miguel~A Munoz.
\newblock Frustrated hierarchical synchronization and emergent complexity in
  the human connectome network.
\newblock {\em Scientific reports}, 4:5990, 2014.

\bibitem{wei2015hcp}
Yunchao Wei, Wei Xia, Min Lin, Junshi Huang, Bingbing Ni, Jian Dong, Yao Zhao,
  and Shuicheng Yan.
\newblock Hcp: A flexible cnn framework for multi-label image classification.
\newblock {\em IEEE transactions on pattern analysis and machine intelligence},
  38(9):1901--1907, 2015.

\bibitem{xie2019snas}
Sirui Xie, Hehui Zheng, Chunxiao Liu, and Liang Lin.
\newblock Snas: stochastic neural architecture search.
\newblock In {\em International Conference on Learning Representations}, 2019.

\bibitem{you2020greedynas}
Shan You, Tao Huang, Mingmin Yang, Fei Wang, Chen Qian, and Changshui Zhang.
\newblock Greedynas: Towards fast one-shot nas with greedy supernet.
\newblock {\em arXiv preprint arXiv:2003.11236}, 2020.

\bibitem{yu2019playing}
Haonan Yu, Sergey Edunov, Yuandong Tian, and Ari~S Morcos.
\newblock Playing the lottery with rewards and multiple languages: lottery
  tickets in rl and nlp.
\newblock {\em arXiv preprint arXiv:1906.02768}, 2019.

\bibitem{zhang2020deeper}
Yuge Zhang, Zejun Lin, Junyang Jiang, Quanlu Zhang, Yujing Wang, Hui Xue, Chen
  Zhang, and Yaming Yang.
\newblock Deeper insights into weight sharing in neural architecture search.
\newblock {\em arXiv preprint arXiv:2001.01431}, 2020.

\bibitem{Zhou2019DeconstructingLT}
Hattie Zhou, Janice Lan, Rosanne Liu, and Jason Yosinski.
\newblock Deconstructing lottery tickets: Zeros, signs, and the supermask.
\newblock In {\em NeurIPS}, 2019.

\bibitem{zhou2016dorefa}
Shuchang Zhou, Yuxin Wu, Zekun Ni, Xinyu Zhou, He~Wen, and Yuheng Zou.
\newblock Dorefa-net: Training low bitwidth convolutional neural networks with
  low bitwidth gradients.
\newblock {\em arXiv preprint arXiv:1606.06160}, 2016.

\bibitem{zoph2016neural}
Barret Zoph and Quoc~V Le.
\newblock Neural architecture search with reinforcement learning.
\newblock {\em arXiv preprint arXiv:1611.01578}, 2016.

\end{thebibliography}
\bibliographystyle{plain}

\appendix

\section{Definitions}
Here is a summary of the notation used throughout this work:

$d \in {\mathbf N}$ maximum number of network connection  

$M_{hard} > 0 $ hard-binarization constant

$0 < M_{soft} \geq M_{hard} $ soft-binarization constant

${\cal X}$ input space

${\cal Y}$ label space

${\cal W} \in \{ \{ 0, 1 \}^d, [0, 1]^d, {\mathbf R}^d \}$ parameter space 

$P_{Z}$ joint object-label distribution 

$Z = (X, Y) \sim P_{Z}$, $X \in {\cal X}$ and $Y \in {\cal Y}$ 
object-label random variable 

${\cal D} = \{ z = (x, y) \text{ realization of } Z \sim P_{Z} \}$ 
training data set

$\psi: {\cal W} \times {\cal X} \to {\cal Y}$ classifier 

$\ell: {\cal Y} \times {\cal Y} \to {\mathbf R}$ loss function 

$f: {\cal W} \times {\cal X} \times {\cal Y} \to {\mathbf R}$ 
single-input classification error

$F: {\cal W} \to {\mathbf R}$,  
$F(w) = |{\cal D}|^{-1} \sum_{z \in {\cal D}} f(w, z) 
\approx {E}_{Z \sim P_Z}(f(w, Z))$ 
average classification error

$B_{M}: {\mathbf R}^d \to [0, 1]^d$, $B_{M}(w) = \sigma_M(w)$ and $M > 0$, binarization function 

$B'_{M}: {\mathbf R}^d \to [0, \frac{M}{4}]^d$, $B'_{M}(w) = \sigma'_M(w)$ 
and $M > 0$, first derivative of the binarization function 

$\sigma_M {\mathbf R}^d \to [0, 1]^d$, $\sigma_M(w) = \frac{1}{1 + e^{-M w}}$ 
and $M > 0$, sigmoid function

$\sigma'_{M}: {\mathbf R}^d \to [0, \frac{M}{4}]^d$, 
$\sigma_M(w) = M \sigma_M(w)\circ (1 - \sigma_M(w))$ and $M > 0$,
first derivative of the sigmoid function 

$[\nabla g(w, z)]_{i} = \frac{\partial g(u, z)}{\partial u_i} |_{u = w}$ 
gradient of $g_z: {\cal W} \to {\mathbf R}$, $g_z(w) = g(w, z)$ at $(w, z) \in {\cal W} \times {\cal X} \times {\cal Y}$ 

$a \in [b, c]$, $a, b, c \in {\mathbf R}^d$ means 
$a_{i} \in [b_i, c_i]$ for all $i=1, \dots, d$

$\text{diag}(a) \in {\mathbf R}^{d \times d}$ is such that 
$[\text{diag}(v)]_{ii} = v_i$ and 
$[\text{diag}(v)]_{ij} = 0$ if $i \neq j$, $i,j = 1, \dots, d$

\section{Assumptions and proofs}

\begin{assumption}
	\label{a assumptions}
	$f:[0, 1]^d \to {\mathbf R}$, 
	is differentiable over $[0, 1]^d$ and obeys   
	\begin{align}
		\max_{v \in [0, 1]^d, z \in {\cal X} \times {\cal Y}} 
		\| \nabla f(v, z) \|^2 &\leq G^2, \\
	\quad
		f(v, z) - f(v', z) &\geq \nabla f(v', z)^T (v - v'), 
\end{align}
	for all $v, v' \in [0, 1]^d$ and $z \in {\cal X} \times {\cal Y}$.
\end{assumption}

\begin{lemma}
	\label{a lemma big F}
	Let 
	$f:[0, 1]^d \to {\mathbf R}$ satisfy Assumption \ref{a assumptions},
	then $F(v): [0, 1]^d \to {\mathbf R}$  
	aslo obeys 
	\begin{align} 
		F(v) - F(v') \geq \nabla F(v')^T (v - v'), 
	\end{align}
	for all $v, v' \in [0, 1]^d$ and $z \in {\cal X} \times {\cal Y}$.
\end{lemma}

\paragraph{Proof of Lemma \ref{a lemma big F}}
The convexity of $f$ implies the convexity of $F$ as
\begin{align}
	F(v) - F(v')  &= |{\cal D}|^{-1}\sum_{z \in {\cal D}} 
	f(v, z) - f(v', z) \\
	&\geq  
	|{\cal D}|^{-1}\sum_{z \in {\cal D}}\nabla f(v', z)^T (v - v') \\
	&= \nabla F(v')^T (v - v').
\end{align}
$\square$

\begin{lemma}
	\label{a lemma error}
	    Let $\eta_t > 0$ , $t = 1, \dots, T$, and 
	    $w^t \in {\cal W}$ be defined by 
	    \begin{align}
		    \label{a proof updates w}
		    w^{t+1} = w^t 
		    - \eta_t \nabla f(B_{M_{hard}}(w^t), z) 
		    \circ B'_{M_{soft}}(w^t), 
		    \quad  t = 1, \dots, T, 
	    \end{align}
	    and for any $z \in {\cal D}$.
	    Then $v^t = B_{M_{hard}}(w^t)$ obey
	   \begin{align}
		\label{a proof updates v}
		v^{t+1}  = v^t - \eta_t (\nabla f(v^t, z) + r^t), 
		    \quad  t = 1, \dots, T, 
    	\end{align}
    	where 
    	\begin{align}
    	\label{a proof r}
		r^{t} &= \nabla f(v^t, z)  
			- B_{M_{hard}}'(\xi^t) 
			\circ \nabla f(v^t,z) \circ B_{M_{soft}}'(w^t) \\ 
		\xi^t &\in [w^t, w^t - 
		\eta_t \nabla f(v^t, z) \circ B_{M_{soft}}(w^t) ]
	\end{align}
	for all $z \in {\cal Z}$.
	Furthermore, the error terms, $r^t$, $t = 1, \dots, T$, obey 
	\begin{align}
		\| r^t \|^2 \leq G^2 C^2 \quad C 
		= 1 + \frac{M_{hard} M_{soft}}{16} 
	\end{align}
	with $G$ defined in (\ref{a assumptions}).
\end{lemma}

\paragraph{Proof of Lemma\ref{a lemma error}}
Let $v = \sigma_M(w)$, for any $w \in {\mathbf R}^d$ and $M > 0$.
Then \eqref{a proof updates w} is equivalent 
to \eqref{a proof updates v} as 
\begin{align}
	v^{t + 1} &= \sigma_{M_{hard}}\left(w^{t} - 
	\eta_t \nabla f(\sigma_{M_{hard}}(w^t), z) 
	 \circ \sigma_{M_{soft}}'(w^t)\right) \\
	 &=  v^{t} 
	 - \eta_t \sigma_{M_{hard}}'(\xi^t) \circ 
	 \nabla f(\sigma_{M_{hard}}(w^t), z) 
	 \circ \sigma_{M_{soft}}'(w^t) \\
	 & = v^{t} - \eta_t \nabla f(v^t, z) + 
	 \eta_t r^t \\
	 r^t &= \nabla f(v^t, z) - 
	 \text{diag}\left(\sigma_{M_{hard}}'(\xi^t) 
	 \circ \sigma_{M_{soft}}'(w^t) \right) 
	 \cdot \nabla f(\sigma_{M_{hard}}(w^t), z) \\
	 & = \nabla f(v^t, z) - 
	 \text{diag}\left(\sigma_{M_{hard}}'(\xi^t) 
	 \circ \sigma_{M_{soft}}'(w^t) \right) 
	 \cdot \nabla f(v^t, z)\\
	 \xi^t &\in [ w^t, w^t - \nabla f(v^t,z) 
	 \circ \sigma_{M_{soft}}'(w^t)]
\end{align}
where the first equality follows from the mean value theorem
For any $w \in {\mathbf R}$ and $z \in {\cal Z}$, 
one has 
\begin{align}
	\label{a proof bound error}
	\| r^t \|^2  &= \| \nabla f(v^t, z)\|^2 
	+ \| \text{diag}\left(\sigma_{M_{hard}}'(\xi^t) 
	\circ \sigma_{M_{soft}}'(w^t) \right) 
	 \cdot \nabla f(v^t, z) \|^2 \\ 
	  & \quad - 2 \nabla f(v^t,z)^T  
	  \text{diag}\left(\sigma_{M_{hard}}'(\xi^t) 
	  \circ \sigma_{M_{soft}}'(w^t) \right) 
	 \cdot \nabla f(v^t, z) \\
	 &\leq G^2 + G^2 \left(\frac{M_{soft} M_{hard}}{16}\right)^2 
	 + 2 G^2 \frac{M_{soft} M_{hard}}{16}  \\
	 & = G^2 \left(1 + \frac{M_{soft} M_{hard}}{16}\right)^2
\end{align}
as, by definition, 
$G^2 = \max_{(v, z) \in [0, 1]^d \times {\cal x}\times {\cal Y}} 
\| \nabla f(v, z)\|^2 $ and 
we have used 
$\max_{w \in {\mathbf R}^d} \sigma_M'(w) = \frac{M}{4}$ for 
any $M > 0$, 
the Cauchy-Schwarz inequality $v^T v' \leq \|v\| \| v' \|$ and 
$v^T \text{diag}(v') \cdot v \leq \max_i v'_i \|v\|^2$.
$\square$

\begin{theorem}
	\label{a theorem convergence}
	Let $f: [0, 1]^d \to {\mathbf R}$ satisfy Assumption 
	\eqref{a assumptions},  
	$\{ w^t \in {\mathbf R}^d \}_{t = 1}^T$
	be the sequence of weights defined in \eqref{a proof updates w}, 
	$\eta_t = \frac{c}{t}$, $t = 1, \dots, T$, and $c > 0$. 
	Then 
	\begin{align}
		{E}\left(
		F(B_{M_{hard}}(w^T)) - F(B_{M_{hard}}(w^*))
		\right) 
		\leq 
		G^2 \frac{c \left(1 + C^2\right)}{2} \frac{1 + \log T}{T},
	\end{align}
	where ${E}(w, z)$ is a short notation for 
	${E}_{Z \sim P_{Z}}(w, Z)$ and 
	$G^2$ and $C$ are defined in Lemma \ref{a lemma error}.
\end{theorem}

\paragraph{Proof of Theorem \ref{a theorem convergence}}
Assumption \eqref{a assumptions} and Lemma \eqref{a lemma big F} 
imply that $F:[0, 1]^d \to {\mathbf R}$ is a convex function 
over $[0, 1]^d$.
The first part of Lemma \ref{a lemma error} implies
that the sequence of approximated ${\mathbf R}^d$-valued  
gradient updates \eqref{a proof updates w}
can be rewritten as the sequence of approximated $[0, 1]^d$-valued 
gradient updates \eqref{a proof updates v}.
The second part of Lemma \ref{a lemma error} implies 
that the norm of all error terms in \eqref{a proof updates v} 
is bounded from above.
In particular, as each $r^t$ is multiplied by the learning rate, $\eta_t$, 
it is possible to show that 
\eqref{a proof updates v} converges 
to a local optimum of $F:[0, 1]^d \to {\mathbf R}$.
This implies that \eqref{a proof updates w}
converges to a local optimum of $F:{\mathbf R}^d \to {\mathbf R}$, 
since, by definition, the mapping $F(B_{hard}(w))$ is 
one-to-one and hence 
\begin{align}
	v^* &:= \text{arg } \min_{v\in [0, 1]^d} F(v)\\  
	&= B_{M_{hard}}\left(
	\text{arg} \min_{w \in {\mathbf R}^d} F(B_{M_{hard}}(w))
	\right)\\
	&=: B_{M_{hard}}(w_*)
\end{align}

To show that \eqref{a proof updates v} converges 
to a local optimum of $F:[0, 1]^d \to {\mathbf R}$
we follows a standard technique for proving the 
convergence of stochastic and make the further (standard) assumption 
\begin{align}
	E(r^t) = 0, \quad t = 1, \dots, T.
\end{align}
First, we let $v^t$, $r^t$ and $\eta_t$, $t = 1, \dots, T$, 
be defined as in Lemma \ref{a lemma error}, and 
$z \in {\cal D}$ be the random sample at iteration $t + 1$.
Then 
\begin{align}
	\|v^{t + 1} - v^{*} \|^2 & = E\left(\|v^{t + 1} - v^{*} \|^2 \right)\\
	& = \|v^{t} - v^{*} \|^2
	- 2 \eta_t E(\nabla f(v^t, \tilde z) - r^t)^T (v^t - v^*)
	\\ \nonumber &\quad 
	+ \eta^2_t E\left(\| \nabla f(v^t, \tilde z) - r^t\|^2\right)\\
	& \leq \|v^{t} - v^{*} \|^2
	- 2 \eta_t E(\nabla f(v^t, \tilde z)^T (v^t - v^*)
	+ \eta^2_t \left(G^2 + E(\|r^t\|^2)\right)\\
	&  \leq  \|v^{t} - v^{*} \|^2
	+ 2 \eta_t (F(v^*) - F(v^t)) + \eta^2_t G^2(1 + C^2) 
\end{align}
where $G^2$ and $C$ are defined in \eqref{a assumptions} 
and \eqref{a proof bound error}.
Rearranging terms one obtains
\begin{align}
	F(v^t) - F(v^*) & = 
	\frac{\|v^{t} - v^{*} \|^2 - \|v^{t + 1} - v^{*} \|^2}{2 \eta_t}
	+ \frac{\eta_t}{2} G^2(1 + C^2) 
\end{align}
and hence 
\begin{align}
	E\left( F(v^T) - F(v^*)\right) &  =  
	E\left( \frac{1}{T} \sum_{t=1}^T 
	\left( F(v^t) - F(v^*)\right)\right)\\
	&\leq \frac{1}{T} \sum_{t=1}^T E\left( F(v^t) - F(v^*)\right)\\ 
	&= \frac{1}{T} \sum_{t=1}^T \left(
	\frac{\|v^{t} - v^{*} \|^2 - \|v^{t + 1} - v^{*} \|^2}{2 \eta_t} 
	+ \frac{cG^2(1 + C^2)}{2 t}  \right) \\
	& = 	- \frac{\|v^{T + 1} - v^{*} \|^2}{2 \eta_T} 
	+ \frac{c G^2(1 + C^2)}{2 T}  \sum_{t=1}^T \frac{1}{t} \\
	& \leq \frac{cG^2(1 + C^2) }{2} \frac{1 + \log T}{T}
\end{align}
where the second line follows from the Jensen's inequality 
and we use $\sum_{t=1}^T \frac{1}{t} \leq 1 + \log T$.
$\square$

\newpage
\section{More results}
\subsection{MNIST: single class experiments}
\begin{figure}[h!]
\begin{center}
\begin{tabular}{cc}
	\includegraphics[trim=0 0 0 0, clip=0, width=.9\textwidth]{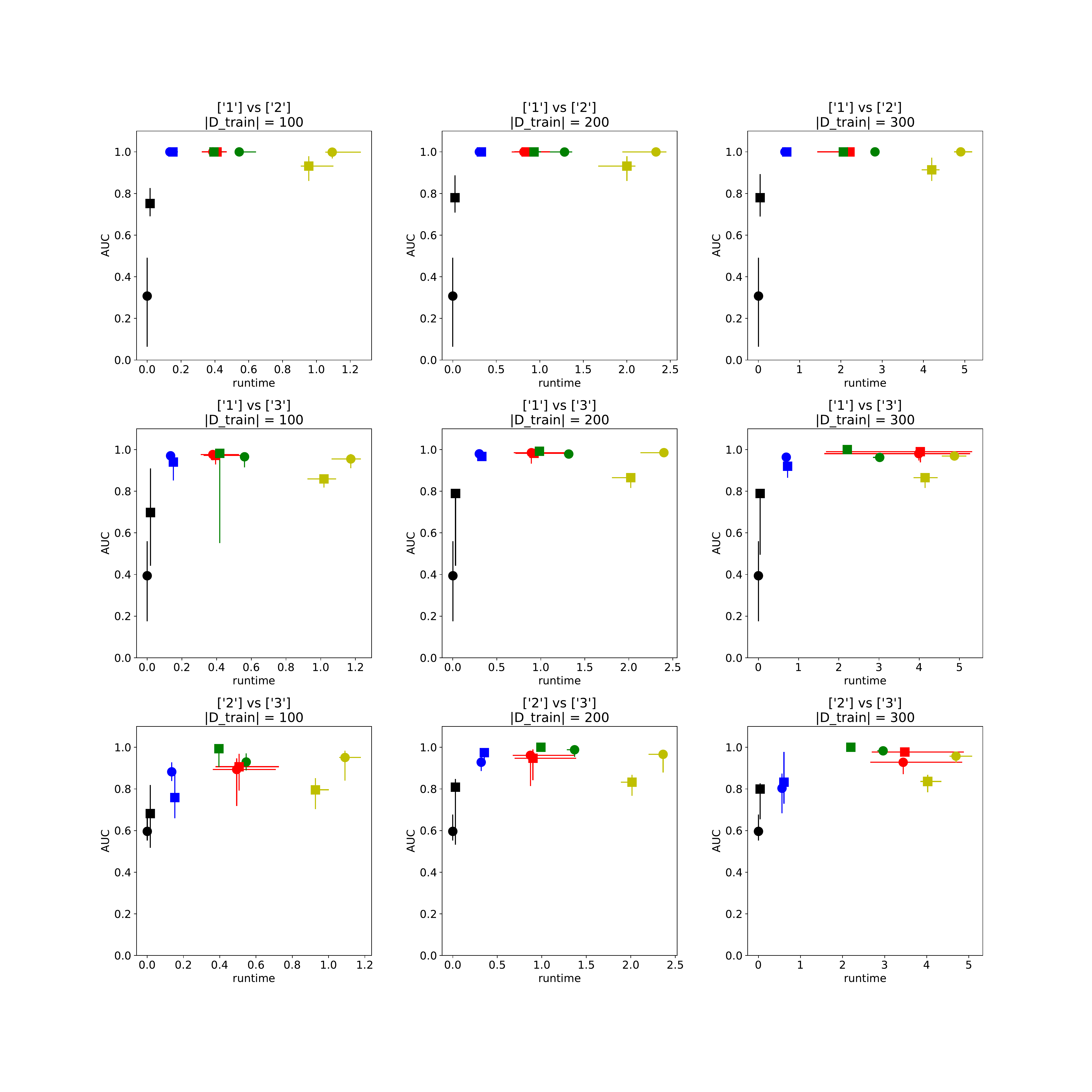} 
	&
	\includegraphics[trim=30 0 0 0, clip=1,width=.1\textwidth]{\imagefolder legend.jpg}
\end{tabular}
	\caption{Runtime and performance of different training strategies 
	and different tasks.
	The size of the training data set and 
	the class names of each experiment 
	are indicated on the title of each plots.
	}
	\label{a mnist single}
\end{center}
\end{figure}
\newpage
\subsection{MNIST: multi-class experiments}
\begin{figure}[h!]
\begin{center}
\begin{tabular}{cc}
	\includegraphics[trim=0 0 0 0, clip=0, width=.9\textwidth]{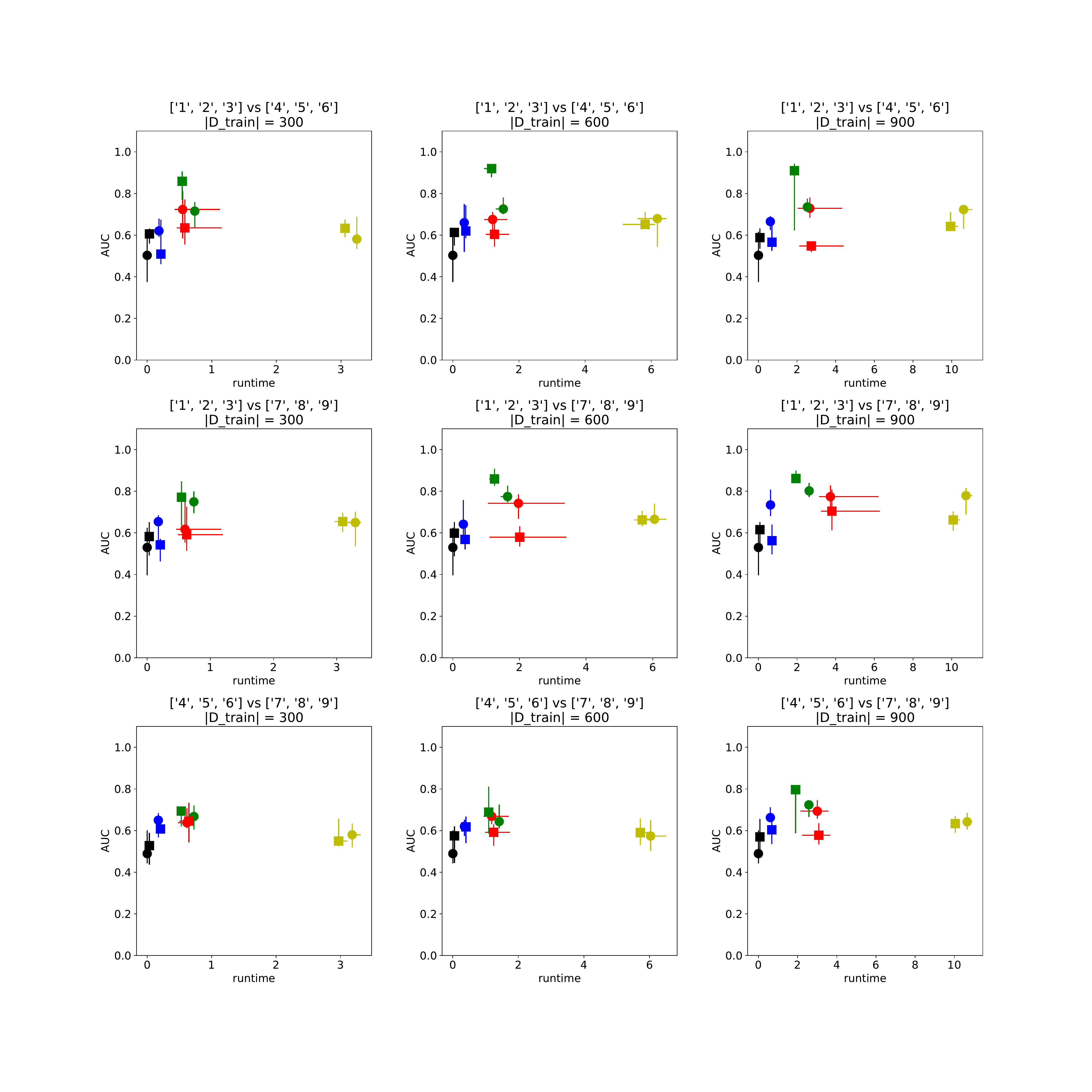} 
	&
	\includegraphics[trim=30 0 0 0, clip=1,width=.1\textwidth]{\imagefolder legend.jpg}
\end{tabular}
	\caption{Runtime and performance of different training strategies 
	and different tasks.
	The size of the training data set and 
	the class names of each experiment 
	are indicated on the title of each plots.
	}
	\label{a mnist multi}
\end{center}
\end{figure}

\newpage
\subsection{Citeseer: single-class experiments}

	\begin{figure}[h!]
\begin{center}
\begin{tabular}{cc}
	\includegraphics[trim=0 0 0 0, clip=0, width=.9\textwidth]{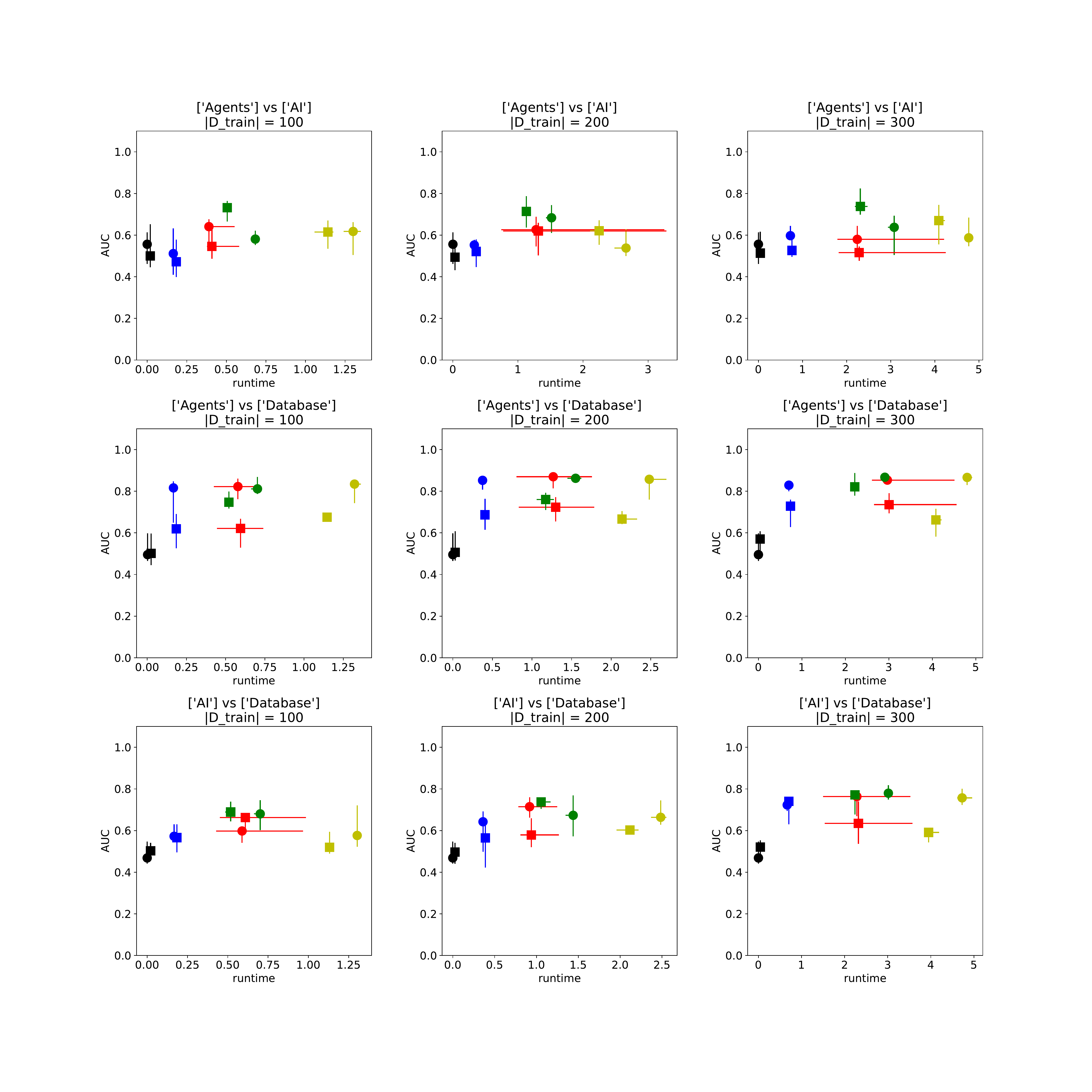} 
	&
	\includegraphics[trim=30 0 0 0, clip=1,width=.1\textwidth]{\imagefolder legend.jpg}
\end{tabular}
	\caption{Runtime and performance of different training strategies 
	and different tasks.
	The size of the training data set and 
	the class names of each experiment 
	are indicated on the title of each plots.
	}
	\label{a citeseer single}
\end{center}
\end{figure}
\newpage
\subsection{Citeseer: multi-class experiments}
\begin{figure}[h!]
\begin{center}
\begin{tabular}{cc}
	\includegraphics[trim=0 0 0 0, clip=0, width=.9\textwidth]{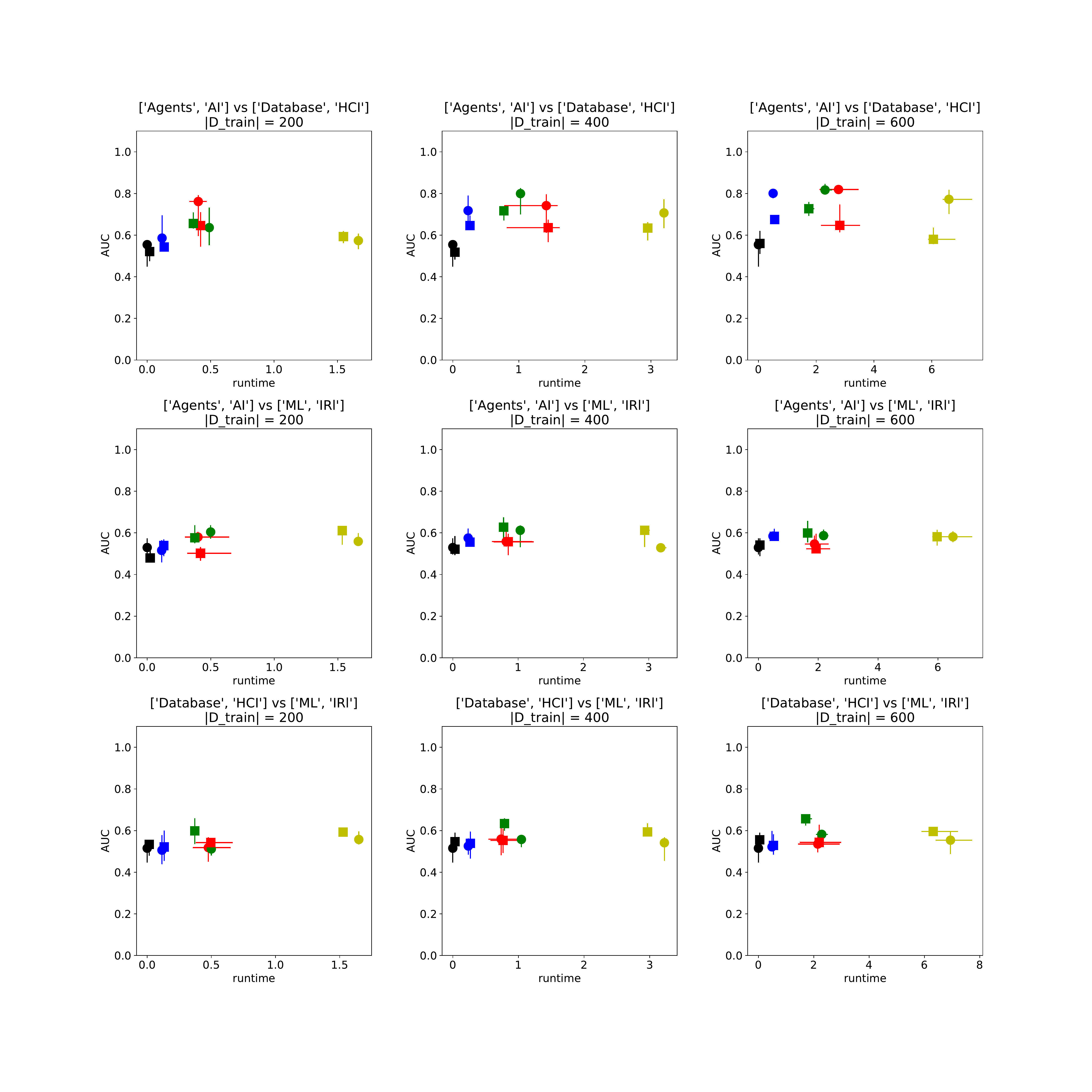} 
	&
	\includegraphics[trim=30 0 0 0, clip=1,width=.1\textwidth]{\imagefolder legend.jpg}
\end{tabular}
	\caption{Runtime and performance of different training strategies 
	and different tasks.
	The size of the training data set and 
	the class names of each experiment 
	are indicated on the title of each plots.
	}
	\label{a citeseer multi}
\end{center}
\end{figure}

\newpage
\section{More Laplacian spectra}
\subsection{Mnist: single-class experiments}
\begin{figure}[h!]
\begin{center}
	\includegraphics[trim=0 0 0 0, clip=0, width=.9\textwidth]{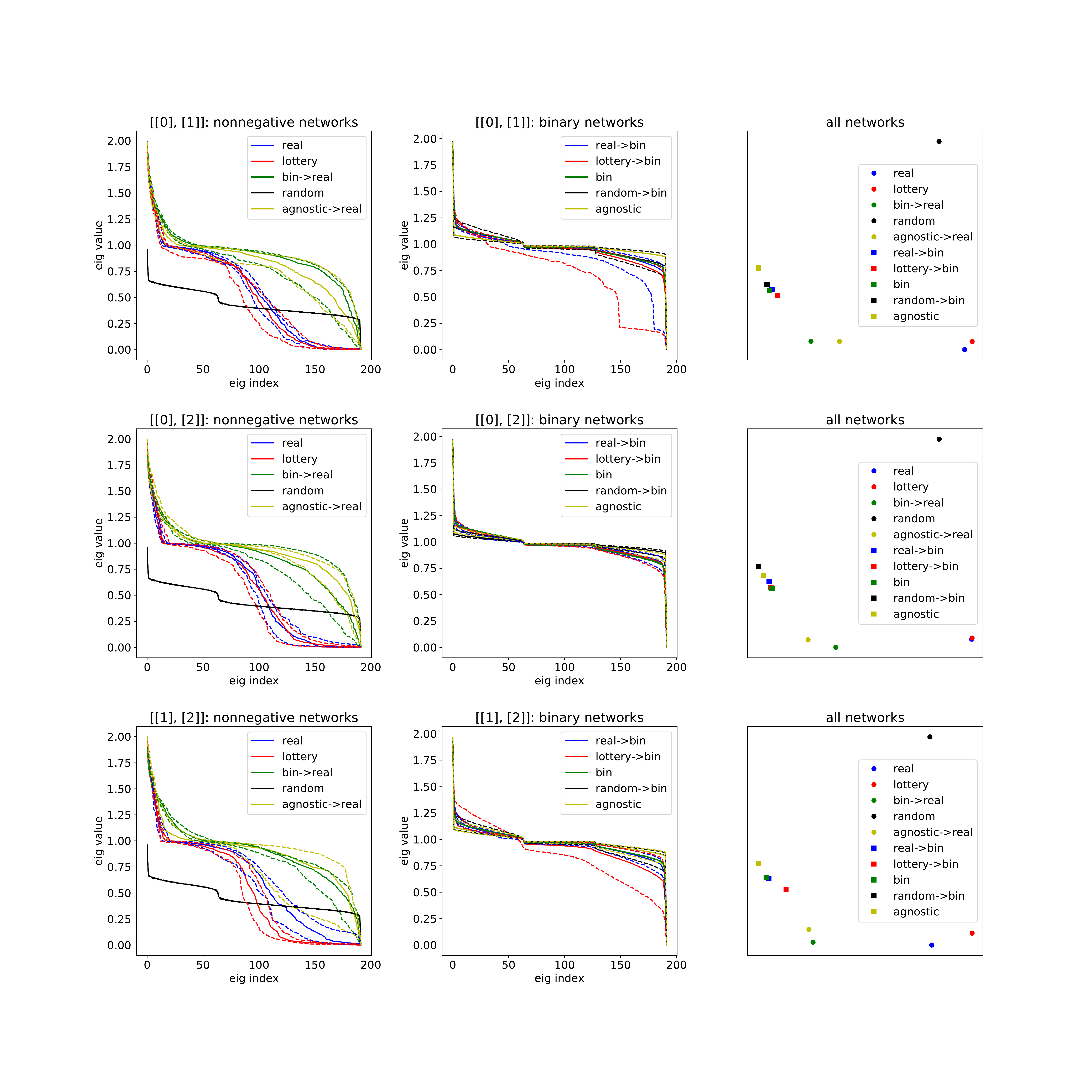} 
	\caption{Spectrum of the normalized Laplacian matrix
	of real and binary 
	models trained on different tasks.
	The size of the training data set is 150 images per class and 
	the class names 
	are indicated on the title of each plots.
	}
	\label{a mnist laplacian}
\end{center}
\end{figure}

\end{document}